\documentclass[11pt]{article}

\usepackage{emnlp2021}
\usepackage{times}
\usepackage{latexsym}
\usepackage[T1]{fontenc}
\usepackage[utf8]{inputenc}
\usepackage{microtype}

\usepackage{array, multirow, bigdelim, makecell, booktabs} 

\usepackage[obeydraft]{todo}
\usepackage[]{graphicx}
\usepackage{booktabs}
\usepackage{makecell}
\usepackage{multirow}
\usepackage{tikz}
\usepackage{pgfplots}
\usepackage{gb4e}
\noautomath

\usetikzlibrary{matrix}
\usetikzlibrary{positioning}

\newcommand{\fyDone}[1]{\done[FY]\Todo[FY:]{\textcolor{violet}{#1}}}
\newcommand{\fyFuture}[1]{\done[FY]\Todo[FY:]{\textcolor{violet}{#1}}}

\usepackage{wasysym}

\renewcommand{\female}{F}
\renewcommand{\male}{M}

\title{Screening Gender Transfer in Neural Machine Translation}

\author{Guillaume Wisniewski \and Lichao Zhu\\
  LLF, Université de Paris \& CNRS  F-75013 Paris, France \\
  \texttt{\{guillaume.wisniewski,lichao.zhu\}@u-paris.fr} \\
  \AND
  Nicolas Ballier \\
  CLILLAC-ARP, Université de Paris\\
  F-75013 Paris, France \\
  \texttt{nicolas.ballier@u-paris.fr} \\
  \And
  Fran\c{c}ois Yvon \\
  Université Paris-Saclay \& CNRS, LISN \\
  91403 Orsay, France \\
  \texttt{francois.yvon@limsi.fr} \\
\\
}

\definecolor{mygrey}{RGB}{229,229,229}
\definecolor{mygrey2}{RGB}{127,127,127}
\definecolor{mygrey3}{RGB}{240,240,240}
 
\pgfplotsset{
        axis background/.style={fill=mygrey},
        tick style=mygrey2,
        tick label style=mygrey2,
        grid=both,
        xtick pos=left,
        ytick pos=left,
        tick style={
                major grid style={style=white,line width=1pt},minor grid style=mygrey3,
                tick align=outside,
        },
        minor tick num=1,
}

\begin{document}
\maketitle

\vspace{\baselineskip}

\begin{abstract}
  This paper aims at identifying the information flow in
  state-of-the-art machine translation systems, taking as
  example the transfer of gender when translating from
  French into English.
  
  Using a controlled set of examples, we experiment several ways to
  investigate how gender information circulates in a encoder-decoder
  architecture considering both probing techniques as well as
  interventions on the internal representations used in the MT system. Our
  results show that gender information can be found in all token
  representations built by the encoder and the decoder and lead us to conclude that 
  there are multiple pathways for gender transfer.\fyDone{there are multiple pathways ?} 
\end{abstract}

\section{Introduction}

The existence of translation divergences (i.e.\ cross-linguistic
distinctions) raises many challenges for machine translation
(MT) \citep{Dorr94machine}: when translating a sentence, some
information or constructions are specific to the target language and,
consequently, can only be inferred by the decoder from the target
context; some are only found in the source language and have to be
ignored; finally, some information has to be adapted and transferred
from the encoder to the decoder. Contrary to previous generations of
MT engines where transfer rules were quite transparent, understanding
this \emph{information flow} within state-of-the-art neural MT systems
is a challenging task, and a key step for their interpretability.

To illustrate these alternatives and the difficulty they raise, we
focus in this work on one specific translation problem: the transfer
of gender information from French, where grammatical gender is a
property of all nouns, and agreement rules exist within the noun
phrase, to English, where gender is only overtly used in rare
constructs involving human agents and pronoun coreference.\footnote{We
  follow \cite{Huddleston02thecambridge} presentation of gender in
  English, mostly grammaticalized in pronouns, as evidenced by
  \textit{herself, himself, itself}.} More specifically, we focus on
the English translation of French sentences such as
``\textit{L'actrice$_{\female}$ a terminé son travail.}'' (the
actress$_{\female}$ 
has finished her$_{\female}$ job).\footnote{In all examples, we tag
  words either with $\female$ or $\male$ to indicate a feminine or
  masculine grammatical gender.} Translating this kind of sentences is
problematic for state-of-the-art MT systems, notably because i) the
coreference has to be correctly identified and ii) it can result in
gender-biased translations due to stereotypical associations such as
\textit{nurses} are always female.

Using a controlled test set, we are able to screen the different information flows
at stake when transferring gender information from French into English using two families of methods. The first one relies
on linguistic probes to find in which parts of the NMT system gender
information is represented; the second one is based on causal models
and consists in intervening on the different parts of the source
sentence and of the decoder representations in order to reveal their
impact on the predicted translations. While this work focuses on one
translation phenomenon and on one translation direction only, we
believe that our observations shed a new light on how translation
systems work and that the methods we describe can be used to analyze
other translation divergences.

The rest of the paper is organized as follows. In
section~\ref{sec:corpus}, we first introduce our controlled dataset
and explain how gender is expressed in the two languages. Then, in
Section~\ref{sec:mtsetup} we describe our MT system and evaluate the system outputs
in Section~\ref{sec:eval} its capacity to translate gender
information. To explain these results, we then describe two sets of
experiments that rely on different ways to analyze neural networks: in
Section~\ref{sec:probes}, we use linguistic probes to find out in
which source and target tokens gender information is present and
in Section~\ref{sec:maniprep} we report experiments modifying the token
representation to determine when this information is used. Finally,
In Section~\ref{sec:related}, we relate our findings to previous
research before summarizing our results in Section~\ref{sec:ccl}.

\section{A Controlled Test Set to Study Gender Transfer between French and English \label{sec:corpus}}

\paragraph{Corpus Creation} Following
\cite{Saunders20genderbias},\footnote{Using a simplified list from
  \cite{Prates19assessing}, \citet{Saunders20genderbias} created a
  ``handcrafted'' dataset of 388 parallel sentences of the type
  \textit{The [PROFESSION] finished [his|her] work.} for three
  translation directions (English-Spanish, English-German and
  English-Hebrew). In this paper, we adapted this approach for a new
  translation direction (French to English) using a much larger list
  of occupational nouns: our corpus contains 3,394 sentences.} we
consider parallel sentences with the following pattern to study gender
transfer between French and English:
\begin{exe}
\ex \texttt{[DET]} \texttt{[N]} a terminé son travail.
\ex The \texttt{[N]} has finished \texttt{[PRO]} work.
\end{exe}
where \texttt{N} is a job noun that can be either masculine or
feminine (e.g.\ in English, actor$_\male$/actress$_\female$; in
French, \textit{acteur}$_\male$, \textit{actrice}$_\female$),
\texttt{DET} is the French determiner in agreement with the noun
(either the feminine form \textit{la$_\female$}, the masculine form
\textit{le$_\male$} or the epicene form \textit{l'}\footnote{The
  French determiner is \textit{l'} for both genres if the job noun begins with a
  vowel.}) and \texttt{PRO} is the English possessive pronoun `\textsl{her}' or `\textsl{his}'.

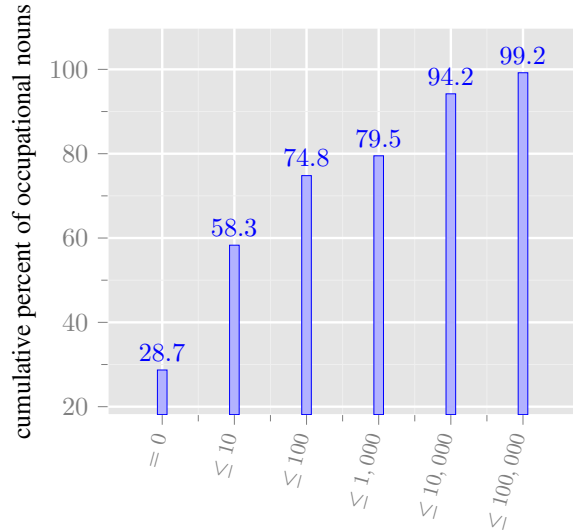
\begin{figure}[htbp]
  \centering
  \begin{tikzpicture}[scale=0.9]
    \begin{axis}[
      draw=white,
      ybar=0pt,
      bar width=4pt,
      enlargelimits=0.15,
      ylabel={cumulative percent of occupational nouns},
      x tick label style={font=\footnotesize,rotate=75,anchor=east,},
      xtick=data,
      nodes near coords,
      nodes near coords align={vertical},
      xticklabels={
        {$=0$},
        {$\leq 10$},
        {$\leq 100$},
        {$\leq 1,000$},
        {$\leq 10,000$},
        {$\leq 100,000$}
        },
      ]
      \addplot coordinates {
        (0,28.7) 
        (1,58.3) 
        (2,74.8) 
        (3,79.5) 
        (4,94.2) 
        (5,99.2)};
    \end{axis}
  \end{tikzpicture}
  \caption{Cumulative frequency of occupational nouns in the training data. \label{fig:train_freq}}
\end{figure}

\begin{figure}[!htpb]
\centering
\begin{tikzpicture}[scale=0.9]
\begin{axis}[
    draw=white,
    ybar=0pt,
    bar width=4pt,
    enlargelimits=0.15,
    ylabel={Counts},
    xlabel={\#BPE token},
    xtick=data,
    nodes near coords,
    nodes near coords align={vertical},
    every node near coord/.append style={rotate=45, anchor=west, font=\footnotesize},
    ]
\addplot coordinates {(1,574) 
                      (2,1185) 
                      (3,1093) 
                      (4,436) 
                      (5,94) 
                      (6,8) 
                      (7,4)};
\end{axis}
\end{tikzpicture}
\caption{Distribution of the number of tokens in French occupational
  nouns after BPE tokenization. \label{fig:bpe_tokens} }
\end{figure}
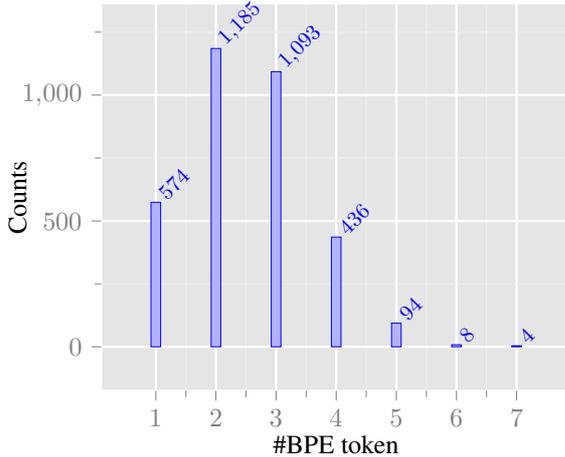

We use the complete list of professions and occupations for French
from \cite{Dister14mettre} to fill the French \texttt{[N]} slot, and
select the associated determiner accordingly. This list contains the
feminine and masculine forms of each profession, allowing us to create
a list of 3,394~sentences, perfectly balanced between
genders.\footnote{This dataset can be found at
  \url{https://github.com/neuroviz/neuroviz/tree/main/blackbox2021}.}
Most of these occupational nouns are rare compound nouns unseen in the
training corpus: as reported in Figure~\ref{fig:train_freq}, only
1,707 of the 2,393 occupational nouns used to create the corpus can be
found in the training set. This is also reflected in
Figure~\ref{fig:bpe_tokens}, where we see that most occupational nouns
are tokenized into multiple~BPE units.  These sentences were
automatically translated and manually verified to produce the
corresponding English list.

The motivations for using these fixed syntactic patterns are
many. First, they limit the only source of variability between
sentences to the \texttt{[N]} slots, allowing us to perform controlled
experiments. Second, they simplify the analysis and manipulation of
the trained representations, as the position of each word is almost
constant throughout the entire dataset.\footnote{Notwithstanding small
  variations due to the BPE segmentation of the noun \texttt{N}, which
  can be split into one, two, three, and up to seven subword units.  }
Despite its simplicity, this dataset enables rich analyses, as the
large coverage of the set of nouns enables us to analyze the result
with respect to the noun frequency, length, stereotypicality, and also
with respect to the amount of gender information available in each
language. Furthermore, the asymmetry between the gender carrying words
in French and English will be a facilitating factor for generating
interesting contrasts. In this paper we focus on French to English
translations,\footnote{For a preliminary analysis of the English to
  French direction, see \citep{Wisniewski21biais}.} using variations
in French determiners to generate interesting contrasts in the source.
The possessive marker \textit{son} is similarly epicene if the
following word begins with a vowel.  It should also be noted that
there are sociolinguistic implications beyond the remit of this paper,
such as a debated preference to refer to women's occupations favouring
the masculine form of the occupational noun or exclusive uses of the
masculine form to express generic uses
\cite{Brauer08ministre}.\fyDone{Missing reference}

\paragraph{Expression of Gender} 

When translating from French, four situations can occur:
\begin{itemize}
\item[(i)] gender information can be inferred from both determiner
  \texttt{DET} and noun \texttt{N}, as in \textit{le$_\male$
    couturier$_\male$}/\textit{la$_\female$ couturière$_\female$} ---
  both translated by `the stylist' or `the seamstress'
  ;
\item[(ii)] gender information can be inferred from the determiner
  \texttt{DET} but not from the noun \texttt{N}, as in
  \textit{le$_\male$ cinéaste}/\textit{la$_\female$ cinéaste} --- `the
  film-maker' in both cases;
\item[(iii)] gender information can be inferred from the noun \texttt{N}
  but not from the determiner \texttt{DET}, as in
  \textit{l'assistant$_\male$}/\textit{l'assistante$_\female$} --- `the
  assistant' in both cases;
\item[(iv)] gender information can not be inferred at all, as in
  \textit{l'illusioniste} --- `the illusionist'.
\end{itemize}
Contrary to case (iv), in situations [i-iii], the translation system has
the information to predict the right pronoun. Table~\ref{tab:n_cases}
reports the number of sentences for each of these cases.

\begin{table}
  \begin{tabular}{lllr}
    \toprule
    determiner & job gender & case & \#sentences     \\
    \midrule
    l' & fem.    &  (iii) & 251 \\
       & epicene &  (iv)  & 272 \\
       & masc.   &  (iii) & 251 \\
    la & fem.    &  (i)   & 895 \\
       & epicene &  (ii)  & 415 \\
    le & epicene &  (ii)  & 417 \\
       & masc.   &  (i)   & 893 \\
    \bottomrule
  \end{tabular}
  \centering
  \caption{Number of sentences for each way of expressing gender in
    French sentences. \label{tab:n_cases}}
\end{table}

Conversely, in English sentences, gender information is always overtly expressed in the English pronoun, and in rare cases, also in
the English noun \texttt{[N]}, as in the \textsl{actor/actress} pair
of words.\footnote{Note that two English sentences (one with `\textsl{her}' the
  other with `\textsl{him}') were created when the gender of the profession can not be inferred from the French sentence. For instance, the French
  sentence `\textit{l'artiste a terminé son travail}' appears twice in
  the parallel corpora: the first time as the translation of `the
  artist has finished \emph{her}$_{\female}$ work', the second time as
  the translation of `the artist has finished \emph{his}$_{\male}$
  work'.}  


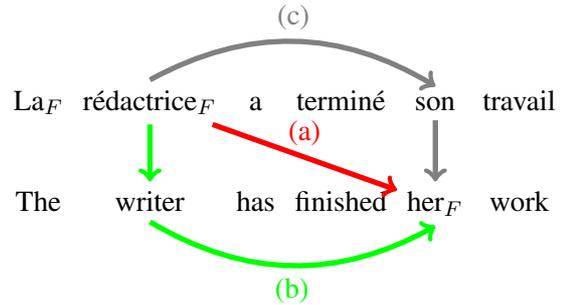
\begin{figure}
    \begin{tikzpicture}[ampersand replacement=\&]
    \matrix[matrix of nodes, row sep=2em] (m) {
      La$_{\female}$ \& rédactrice$_{\female}$ \& a \& terminé \& son \& travail \\
      The \& writer \& has \& finished \& her$_{\female}$ \& work \\
    };
    
    \draw[->, green, line width=2pt] (m-1-2) -- (m-2-2);
    \draw[->, red, line width=2pt]   (m-1-2) -- (m-2-5) node[above, pos=.5] {(a)};
    \draw[->, gray, line width=2pt]  (m-1-5) -- (m-2-5);
    
    \path (m-1-2.north) edge[->, line width=2pt, gray, bend left] node[above] {(c)} (m-1-5.north);
    \path (m-2-2.south) edge[->, line width=2pt, green, bend left=-30] node[below] {(b)} (m-2-5.south);
  \end{tikzpicture}
  \centering
  \caption{%
    Gender transfer, from French to English: three possible influences
    on the choice of the gender of the possessive pronoun in
    English.
    \label{fig:gendertrans}
  }%
\end{figure}

\paragraph{Pathways to Transfer Gender Information} Looking at the
example in Figure~\ref{fig:gendertrans}, we see that to correctly
translate the gender of the French profession into English, three main
hypotheses can be entertained:
\begin {itemize}
\item (a) a \emph{direct influence} through the cross-lingual
  attention(s) computed when generating the English pronoun that
  should attend to the French noun;
\item (b) an \emph{indirect influence} through the (monolingual)
  encoding of gender in the representation of the English noun, the
  contextualized embedding of which should encode (through
  cross-lingual projection) the gender of the corresponding
  French NP;
\item (c) an \emph{indirect influence} through the (cross-lingual)
  attention to the French possessive adjective, the contextualized
  representation of which should then encode the French Noun gender.
\end{itemize} 
Note that these three possibilities are not mutually exclusive, and
gender may well be transferred through a combination of the three
influences, and also through the representations computed for the
other words in the sentence.

Our main objective in this paper is to explore various ways to assess
these hypotheses and try to reach a conclusion regarding
the way gender is actually transferred.

\section{Experimental Setting\label{sec:mtsetup}}

In all our experiments, we use
\texttt{JoeyNMT}\footnote{\url{https://github.com/joeynmt/joeynmt}}
\citep{Kreutzer19joey}, an educational implementation of a
translation system based on the \texttt{Transformer} model of
\citet{Vaswani17attention}. The simplicity of the codebase, which
nonetheless allowed us to achieve near SOTA performance on our data,
made it a perfect choice for our endeavor. In our system, encoder and
decoder are composed of 6~layers, each with 8~attention heads; the
\textit{feed-forward} layers have 2,048 parameters and the dimension
of lexical embeddings is~512. Our model comprises a grand total of
76,596,736 parameters. The system was trained with data from the
`News' task of the WMT'2015 evaluation campaign.\footnote{This is the
  most recent evaluation campaign for English-French organized in the
  context of the WMT conference (see
  \url{http://statmt.org/wmt15}).} It includes the
\texttt{Europarl}, \texttt{NewsCommentary} and \texttt{CommonCrawl}
corpora, and altogether contains 4,813,682 sentences and nearly 141
million French running words. All the corpora were tokenized and
segmented into sub-lexical units using the unigram model of
\texttt{SentencePiece} \citep{Kudo18subword}; the resulting
vocabularies contain 32,000 units in each language. The model is
trained by optimizing the cross-entropy using the \textsc{Adam}
strategy. This system achieves a BLEU score of 34.0 for the
French-English direction.

\section{Evaluation of Gender Translation \label{sec:eval}}

\subsection{Experimental Results}

We evaluate the ability of our system to predict the gender of 
occupational nouns using the corpus described in 
Section~\ref{sec:corpus} and consider, as a point of comparison, the
translations generated by \texttt{e-translation}, a translation system
developed by the European Commission that is freely accessible for
academic research.\footnote{\url{ec.europa.eu/cefdigital/eTranslation}} When
translating into English, this evaluation is straightforward and
simply amounts to checking the pronoun gender: does the translation
hypothesis of a feminine (resp.\ masculine) occupational noun contains \textit{her}
(resp. \textit{his})? We therefore evaluate the two considered systems
by the percentage of sentences for which the possessive pronoun is
correct.


It should be noted that the gender information of a translation
hypothesis can not always be determined: in some cases, the system
produces a correct translation that does not contain \textit{her} nor
\textit{his} (e.g.\ \textit{the programmer has finished working}); in
other cases, the translation is completely wrong or the determiner is
translated as \textit{its} (901~sentences mostly corresponding to
situations in which the job noun was not translated correctly) or as
\textit{their} (52~sentences). For the sake of clarity, we do not
distinguish these cases in our analyses.

It appears that our system is able to correctly predict the possessive
pronoun in only 52.4\% of the English sentences (the gender
information could not be extracted in 1.4\% of the sentences); on the
contrary, \texttt{e-translation} achieves near perfect results: in
90.9\% of the translation hypotheses, the gender of the pronoun is
correct, which strongly suggests that this system integrates a
specific process to transfer gender information.

Table~\ref{tab:res_gender_prediction} details these scores for the
various situations identified in Section~\ref{sec:corpus}. These
results show that our system (trained on `standard' MT corpora) has a
clear tendency to favor the translation of \textit{son} by a masculine
pronoun even in situations in which there is no ambiguity on the
gender of the nominal group (e.g.\ when both the determiner and the
noun both have a form specific to the feminine). Overall, our system
achieves a precision of only 26.3\% for the feminine pronouns, but
correctly predicts the pronoun for 78.5\% of masculine sentences. These
observations are in line with the conclusions drawn by
\newcite{Saunders20genderbias} on English-German, English-Spanish and
English-Hebrew. Similar observations are also reported in \citep{Renduchintala21investigating} when translating out of English for a larger set of target languages. On the contrary, \texttt{e-translation} is able to
correctly infer the gender information in almost all cases and most of
the errors are due to the French sentences in which the gender is not
expressed (case (iv) in the description of Section~\ref{sec:corpus}).

\begin{table*}
  {\footnotesize
  \begin{tabular}{l c l l rl l rl}
    \toprule
               &            &              &   & \multicolumn{2}{c}{\texttt{JoeyNMT}} & \phantom{abc} & \multicolumn{2}{c}{\texttt{e-translation}} \\
    \cline{5-6} \cline{8-9}
    Determiner & Job gender & \makecell{predicted \\ pronoun} & \phantom{abc} & \% sentences & accuracy & & \% sentences & accuracy\\ 
    \midrule                    
    l'         & epicene    & her          &    &  0.7\% & & & 4.4\% \\
               &            & his          &    & 80.1\% & & & 94.1\% \\
               &            & other        &    & 19.2\% & \rdelim\}{-3}{*}[40.4\%] & & 1.5\% & \rdelim\}{-3}{*}[49.3\%] \\
               & fem.       & \textbf{her} &    &  7.2\% & & & 91.6\% \\
               &            & his          &    & 59.4\% & & &  4.0\% \\
               &            & other        &    & 33.4\% & \rdelim\}{-3}{*}[7.2\%] & & 4.4\% & \rdelim\}{-3}{*}[91.6\%] \\
               & masc.      & her          &    &  0.4\% & & & 0\% \\
               &            & \textbf{his} &    & 73.7\% & & & 96\% \\
               &            & other        &    & 25.9\% & \rdelim\}{-3}{*}[73.7\%] & & 4.0\% & \rdelim\}{-3}{*}[96.0\%] \\
    la         & epicene    & \textbf{her} &    & 31.6\% & & & 93\% \\
               &            & his          &    & 43.9\% & & & 0.3\% \\
               &            & other        &    & 24.5\% & \rdelim\}{-3}{*}[31.6\%] & & 6.7\% & \rdelim\}{-3}{*}[93.0\%] \\
               & fem.       & \textbf{her} &    & 33.3\% & & & 94\% \\
               &            & his          &    & 18.5\% & & & 0\% \\
               &            & other        &    & 48.2\% & \rdelim\}{-3}{*}[33.3\%] & & 6\%  & \rdelim\}{-3}{*}[94.0\%] \\
    le         & epicene    & her          &    &  0.7\% & & & 0\% \\
               &            & \textbf{his} &    & 84.4\% & & & 95.4\% \\
               &            & other        &    & 14.9\% & \rdelim\}{-3}{*}[84.4\%] & & 4.6\% & \rdelim\}{-3}{*}[95.4\%] \\
               & masc.      & her          &    &  0.2\% & & & 0\% \\
               &            & \textbf{his} &    & 76.8\% & & & 95.4\% \\
               &            & other        &    & 21.2\% & \rdelim\}{-3}{*}[76.8\%] & & 4.4\% & \rdelim\}{-3}{*}[95.6\%] \\
    \bottomrule
  \end{tabular}
  \centering
  \caption{Percentage of translation hypotheses that contain each
    possessive pronoun according to the way gender is expressed in
    the French subject. In each case the correct English pronoun is in
    bold. \label{tab:res_gender_prediction}}
}
\end{table*}

\subsection{Predicting Failure \label{sec:predicting}}


We conducted two experiments to better understand the reasons why the
gender of the possessive pronoun is not correctly predicted in the
translations of \texttt{JoeyNMT}.

First, we looked at the number of times the possessive pronoun
\textit{son} was translated by \textit{his} or \textit{her} in the
training data. For this purpose, we used
\texttt{eflomal}~\cite{Ostling2016efmaral} to align French and English
tokens of the training set\footnote{Alignment was performed after BPE
  tokenization and symmetrized using the \texttt{grow-diag-final-and}
  heuristic.} and use the alignment link to find all possible
translation of the French \textit{son} token.\footnote{Note that, in
  French, \textit{son} can either be a possessive pronoun or a noun
  meaning `sound'.} Results reported in Table~\ref{tab:son_trans} show that
translations of \textit{son} by \textit{his} are three times more
frequent than translations by \textit{her}.

\begin{table}
  \centering
  {\footnotesize
    \begin{tabular}{lc}
      \toprule
      translation & frequency  \\
      \midrule
      \_its     & 27.94\% \\
      \textbf{\_his}     & \textbf{18.28\%} \\
      \_the     &  7.24\% \\
      \textbf{\_her}     &  \textbf{6.42\%} \\
      \_a       &  3.34\% \\
      \_their   &  2.92\% \\
      \_it      &  2.45\% \\
      \_sound   &  1.37\% \\
      s         &  1.33\% \\
      \_he      &  0.76\% \\
      \midrule
      \texttt{\_\_OTHER\_\_} & 22.13\% \\
      \bottomrule
    \end{tabular}
    \caption{Most frequent translation of the French token
      \textit{son} according to the word alignment
      links. \textit{son} is aligned with 3,658 different types. Those
      which do not appear in the table are grouped in the special
      token \texttt{\_\_OTHER\_\_}. \label{tab:son_trans}}
  }
\end{table}

Second, we considered a simple logistic regression model that, given a
sentence, predicts whether the pronoun gender will be correct or
not. We used a small set of surface features to describe a French sentence: the
gender of the determiner, the gender of the occupational noun (both
can be either masculine, feminine or epicene), a binary feature that
is true when both the determiner and the noun have an explicit gender
marker, a feature describing the number of BPE units into which the
occupational noun has been encoded and three Boolean features to
describe the number of occurrences of the occupational noun in the
train set. These features are respectively true when \fyFuture{The
  logic: 1 time or more} the occupational noun does not appear in the
training set, when it occurs 10 times or less in the training set and
when it occurs 100 times or more in the training set.

This model is trained on 75\% of the examples and we evaluate the
accuracy of its predictions on the remaining 25\% of examples. To
assess the stability of the model we consider 100 train-test splits
and report the 95\% confidence interval.

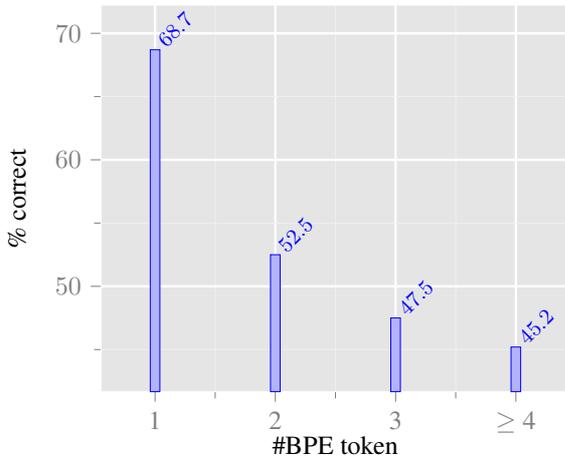
\begin{figure}[htbp]
\begin{tikzpicture}[scale=0.9]
\begin{axis}[
    draw=white,
    ybar=0pt,
    bar width=4pt,
    enlargelimits=0.15,
    ylabel={\% correct},
    xlabel={\#BPE token},
    xticklabels={1,2,3,$\geq$ 4},
    xtick=data,
    nodes near coords,
    nodes near coords align={vertical},
    every node near coord/.append style={rotate=45, anchor=west, font=\footnotesize},
    ]
\addplot coordinates {(1,68.7) 
                      (2,52.5) 
                      (3,47.5) 
                      (4,45.2)};
\end{axis}
\end{tikzpicture}
\caption{Sentences in which the English possessive pronoun is
  correctly translated according the number of tokens in the French occupational nouns after BPE
  tokenization. \label{fig:res_by_bpe} }
\end{figure}

We report in Table~\ref{tab:res_regression} the accuracy achieved
using all of these features as well as each of this feature
individually.\footnote{Regression models have been computed using
  \texttt{sklearn} \cite{scikit-learn}.} Results show that, overall,
the quality of the prediction is pretty high even when considering a
single feature. This observation suggests the choice of the possessive
pronoun in English is mostly based on surface information and does not
result from a `linguistic' analysis of the input sentence. In
particular, the high precision achieved when considering solely the
number of BPE tokens, corroborated by the observation reported in
Figure~\ref{fig:res_by_bpe}, shows that the system is not able to
correctly predict gender information for occupational nouns that it
did not see during training. As expected, the best feature to predict
whether the gender of the English pronoun will be correct is the
combination between the gender of the determiner and the gender of the
job name, a feature that is closely related to the different ways gender is expressed in the French sentence as described in
Section~\ref{sec:corpus}.


\begin{table}
  \begin{tabular}{llr}    
    \toprule
    \multicolumn{2}{l}{feature} & accuracy \\
    \midrule
    \multicolumn{3}{l}{\textit{single features}} \\
                                & occupational noun gender & $73.9\% \pm 0.2$ \\
                                & determiner gender        & $74.5\% \pm 0.2$ \\
                                & number of BPE tokens      & $74.5\% \pm 0.2$ \\
                                & explicit gender          & $74.5\% \pm 0.2$ \\
                                & occurrences in train set & $74.5\% \pm 0.2$ \\
    \midrule
    \multicolumn{3}{l}{\textit{combination of features}} \\
                                & all gender features & $81.5\% \pm 0.2$ \\
                                & all features & $82.2\% \pm 0.2$ \\
    \bottomrule
  \end{tabular}
  \centering
  \caption{Accuracy achieved when predicting whether the possessive
    pronoun will be correctly translated. Features are described in
    Section~\ref{sec:predicting}; `all gender features' denotes the
    combination of three features: occupational noun gender,
    determiner gender and explicit gender. \label{tab:res_regression}}
\end{table}

\section{Probing Representations \label{sec:probes}}

\begin{table}
  \begin{tabular}{cccc}
    \toprule
    & \phantom{abc} & \multicolumn{2}{c}{decoder} \\
    \cline{3-4}
    layer & & the & all tokens \\
    \midrule
    1 & & $89.5\%$ { \tiny $\pm 0.2$} & $71.6\%$ { \tiny $\pm 0.6$ } \\
    2 & & $92.0\%$ { \tiny $\pm 0.1$} & $76.3\%$ { \tiny $\pm 0.7$ } \\
    3 & & $91.8\%$ { \tiny $\pm 0.1$} & $78.1\%$ { \tiny $\pm 0.6$ } \\
    4 & & $90.9\%$ { \tiny $\pm 0.2$} & $79.1\%$ { \tiny $\pm 0.6$ } \\
    5 & & $89.3\%$ { \tiny $\pm 0.2$} & $82.4\%$ { \tiny $\pm 0.5$ } \\
    6 & & $87.7\%$ { \tiny $\pm 0.2$} & $84.7\%$ { \tiny $\pm 0.3$ } \\
    \bottomrule
  \end{tabular}
  \caption{Precision of a probe predicting the gender of the French
    occupational noun given the decoder
    representation. \label{tab:probing_decoder}}
\end{table}

In this section, we conduct an analysis of the representations
computed by the encoder when translating from French into English. Our
goal is to evaluate how well the gender information spreads through
the transformer network from the initial French occupational noun
(either \texttt{DET}, \texttt{N} or both of them) to the other French
words, as well as to their English counterparts. Following a standard
practice, we use probing~\citep{Belinkov19analysis} to analyze which
words in the source and target sentence convey gender information: a
\textit{probe}~\citep{Alain17understanding} is trained to predict
linguistic properties (here the gender of the French subject) from the
representations of language; achieving high accuracy at this task
implies these properties were encoded in the representation.

\paragraph{Experimental setup} We extract and collect the 512
dimensional hidden representations at the output of each layer of the
encoder for all French lexical tokens following the job noun (i.e.\
\textit{a}, \textit{terminé}, \textit{son}, \textit{travail},
\textit{.} and \texttt{<eos>}), as well as the first token
(\textit{the})\footnote{This is the only word that is always predicted
  correctly.} of the English sentence in all decoder layers. All these
words are frequent enough to correspond to one single BPE unit. We
also consider a probe that is trained on all tokens of the target
sentence (i.e.\ we collect the token representations of all
translation hypotheses and associate each of them to a label
indicating whether occupational name in the French sentence refers to
a woman or a man), as the diversity of the translation structures
makes it impossible to carry out a position-by-position analysis.

For each word, we randomly split our 3,394 sentences between train
(75\%) and test (25\%), and use
\texttt{scikit-learn} \citep{scikit-learn} to learn a logistic
regression model that predicts the gender of the occupational noun using the
hidden representation of one single word. We use $\ell_1$ penalty
to regularize this model. The same data is also used to predict a
random binary labeling: this is to control the capacity of our probing
model \cite{Hewitt19designing}. This experiment is repeated on 100
random train/test splits and 95\% confidence intervals are
computed.

\paragraph{Results}
Table~\ref{tab:probing_encoder} reports the accuracy achieved by our
probes considering the representation of the source tokens as
features. It appears that the representation of \textit{son} (the
translation of the possessive pronoun in French) is not the same when
the occupational noun is masculine as when it is feminine: the
representation of the French possessive pronoun encodes gender
information even if the form of the word does not. It also appears
that this information is more present in the deepest layers of the
encoder: the probe achieves an accuracy of 80\% when representations
from the first layer are considered and of more than 90\% when
representations are extracted from any of the last three layers of the
encoder. This observation confirms that there is actually an
information flow between the possessive pronoun \textit{son} and its
antecedent the French occupational noun, corresponding to the path
denoted~(c) in Figure~\ref{fig:gendertrans}.

More surprisingly, accuracies achieved by the probe when the
representations of other source tokens are considered
are also very high: these accuracies are comparable or only slightly less
than the ones achieved with \textit{son}, showing that the gender
information has an impact on the representations of all source tokens,
even when these tokens have no direct syntactic relations with the
subject phrase.

The results of the probe considering the decoder representations of
`\textit{the}' (Table~\ref{tab:probing_decoder}) show a similar trend: the
gender information is encoded in the representation even if the token
generated by the decoder does not change with this information.
It also appears that the probe is still able to predict the gender of the
French occupational noun with a high accuracy when the representation
of any token predicted by the decoder is considered
as features, showing that, as for the encoder representations, gender
information is encoded in all target tokens, even those for which this
information is useless.

Results for predicting random labels (column `random labels' in
Table~\ref{tab:probing_encoder}) finally show that the information is
actually present in the representations and that the probe is not capturing
spurious correlations in our data \cite{Hewitt19designing}.

\begin{table*}
  {\footnotesize
  \begin{tabular}{ccccccccc}
    \toprule
    & \multicolumn{6}{c}{encoder} & & random labels \\
    \cline{2-7} \cline{9-9}
    layer      & a & terminé & son & travail & . & \texttt{eos} & \phantom{abc} & son \\
    \midrule
    1 & $80.4\%$ { \tiny $\pm 1.1$} &$75.1\%$ { \tiny $\pm 0.3$} &$80.6\%$ { \tiny $\pm 0.3$} &$76.4\%$ { \tiny $\pm 0.6$} &$59.5\%$ { \tiny $\pm 1.0$} &$73.3\%$ { \tiny $\pm 1.0$} & & $45,3\%$ { \tiny $\pm 0.9$} \\
    2 & $85.8\%$ { \tiny $\pm 1.0$} &$80.8\%$ { \tiny $\pm 0.2$} &$81.6\%$ { \tiny $\pm 0.3$} &$78.3\%$ { \tiny $\pm 0.7$} &$87.6\%$ { \tiny $\pm 0.6$} &$88.3\%$ { \tiny $\pm 0.7$} & & $50,7\%$ { \tiny $\pm 0.8$} \\
    3 & $89.5\%$ { \tiny $\pm 0.6$} &$88.2\%$ { \tiny $\pm 0.2$} &$89.2\%$ { \tiny $\pm 0.2$} &$82.0\%$ { \tiny $\pm 1.1$} &$86.5\%$ { \tiny $\pm 1.0$} &$87.6\%$ { \tiny $\pm 0.6$} & & $48,8\%$ { \tiny $\pm 0.9$}\\
    4 & $90.8\%$ { \tiny $\pm 0.4$} &$89.3\%$ { \tiny $\pm 0.2$} &$90.6\%$ { \tiny $\pm 0.2$} &$85.9\%$ { \tiny $\pm 0.9$} &$85.7\%$ { \tiny $\pm 1.0$} &$85.6\%$ { \tiny $\pm 0.7$} & & $48,6\%$ { \tiny $\pm 0.8$}\\
    5 & $90.4\%$ { \tiny $\pm 1.0$} &$89.3\%$ { \tiny $\pm 0.2$} &$90.4\%$ { \tiny $\pm 0.2$} &$85.5\%$ { \tiny $\pm 0.8$} &$86.4\%$ { \tiny $\pm £0.8$} &$85.2\%$ { \tiny $\pm 1.2$} & & $49,6\%$ { \tiny $\pm 0.8$}\\
    6  & $91.0\%$ { \tiny $\pm 0.6$} &$89.3\%$ { \tiny $\pm 0.2$} &$90.0\%$ { \tiny $\pm 0.2$} &$86.0\%$ { \tiny $\pm 1.0$} &$86.4\%$ { \tiny $\pm 1.1$} &$85.1\%$ { \tiny $\pm 0.8$} & & $49,2\%$ { \tiny $\pm 0.8$}\\
    \bottomrule
  \end{tabular}
}
  \centering
  \caption{Precision of a probe predicting the gender of the French
    subject given the encoder
    representations. \label{tab:probing_encoder}}
\end{table*}


\section{Manipulating Representations \label{sec:maniprep}}

The probing experiments described in the previous section show that
gender information is encoded in all tokens representations built by
the encoder and the decoder. However, it is not possible to identify
from these observations if and when this information is used. To
answer this question, the second method we propose to analyze gender
transfer in our MT system relies on an \emph{intervention}. It consists
in replacing the embedding of the French possessive pronoun (i.e. the
\textit{son} token that, intuitively triggers the generation of `\textit{her}'
or `\textit{his}') at the output of the encoder by either a \emph{neutral}
version of this embedding, obtained by averaging the representations
of \textit{son} on the whole test set (it should be borne in mind that the design of our corpus 
ensures that genders are balanced) or a \emph{prototypical} version of
a masculine \textit{son} embedding or a feminine \textit{son}
embedding. These embeddings are extracted from the encoder
representations of these two sentences:
\begin{exe}
  \ex le facteur a terminé son travail. \\
  the postman has finished his work.
  \ex la pharmacienne a terminé son travail. \\
  the pharmacist has finished her work.
\end{exe}
These two sentences were chosen because, in both cases, gender
information is carried by both the determiner and the noun and the
translation of these sentences by our system is correct. After
plugging the chosen representation in the last layer of the encoder
and keeping the representation of the other tokens of the source
sentence unchanged, the rest of the translation proceeds without any
further modification.

Results of this manipulation are in Table~\ref{tab:manipson}: like
in Table~\ref{tab:res_gender_prediction}, we have reported the proportion of translation
hypotheses in which the possessive pronoun is feminine, masculine or
is neither \textit{her} nor \textit{his}.\footnote{As in
  Section~\ref{sec:eval}, this last category includes both sentences
  that cannot be analyzed and those in which the possessive pronoun is
  \textit{its} or \textit{their}.}  Contrary to what was expected,
changing the representation of the French possessive 
\textit{son} has little impact (if any) on the choice of the English
pronoun. These observations suggest that the representations
of \textit{son} built by the MT system are not the only evidence used\fyDone{fixed writing}
during the generation of the translation hypothesis, even if the
results reported in the previous section show that these
representations are particularly relevant for making the correct
prediction. This counter-intuitive result is consistent with several
observations made in the literature: the fact that a `linguistic'
information is encoded in the neural representations does not imply that
it will be used by the neural network (see, for instance, \citep{Belinkov19analysis}).
This suggests that the information flow along the path denoted~(c) in
Figure~\ref{fig:gendertrans} should be small and the choice of the English possessive
pronoun is based on other information than the representation of \textit{son}.

\begin{table}
  \centering \footnotesize
  \begin{tabular}{llc}
    \toprule
    intervention   & English pronoun & \% sentences \\
    \midrule
    none           & her    & 13.4\% \\
                   & his    & 57.1\% \\
                   & other  & 29.5\% \\
    feminine       & her    & 17.3\% \\
                   & his    & 56.8\% \\  
                   & other  & 25.9\% \\
    gender-neutral & her    & 13.2\% \\        
                   & other  & 29.4\% \\        
                   & his    & 57.4\% \\        
    masculine      & her    & 13.8\% \\
                   & other  & 29.2\% \\
                   & his    & 57.0\% \\     
    \bottomrule
  \end{tabular}
  \caption{Intervention on \textit{son} representations: proportion of
    translation hypotheses in which the English possessive pronoun is
    \textit{her}, \textit{his} or neither of these two values,
    depending on the intervention on
    \textit{son}. \label{tab:manipson}}
\end{table}


\section{Related Work \label{sec:related}}

Our work is part of a very active line of research aiming to analyze,
interpret, and evaluate neural networks used in
NLP. \newcite{Belinkov19analysis} present a detailed overview of
these papers and of the different tools and methods that can be used
to uncover the linguistic information represented in the hidden layers
of neural networks. Experiments reported in Section~\ref{sec:probes}
are based on the probing approach of \newcite{Alain17understanding}
and have been used in many works (see \citep{Belinkov19analysis} for
an overview). This approach has also been used in several works to
study the information flow within an encoder-decoder architecture:
for instance, \newcite{Belinkov20onthelinguistic} rely on probes to
find which components of a NMT system encode linguistic information
when translating morphologically rich languages. However, to the best
of our knowledge, this work is the first to use the differences
between gender expression in French and English to get insights into the
inner representations used in NMT systems based on the Transformer
architecture. Experiments reported in Section~\ref{sec:maniprep} are
inspired by causal analysis, a type of analysis that has been used by
\newcite{Vig20investigating} to analyze gender bias in neural
monolingual NLP models.

Several studies have investigated gender bias using dedicated datasets, some
of them presented at the ACL Workshop on Gender Bias in Natural Language
Processing
\citep{Gebnlp19gender,Gebnlp20gender,Gebnlp21gender}. \citet{Savoldi21gender}
synthesizes the studies and datasets on gender bias for translation. In
particular, the controlled test set considered in our work builds on the works
of \citet{Stanovsky19evaluating} and \citet{Saunders20genderbias}, who both
propose challenge test sets to evaluate gender bias in MT systems. The
corresponding datasets consider the translation of occupational nouns with an
anaphoric reference that makes gender explicit: the former contains instances of
difficult translation patterns inspired by the WinoGender dataset of
\citet{Rudinger18genderbias}; similar to our work, the latter contains a smaller
set of simple sentences following a fixed template. Working with a slightly more
varied set of sentence templates chosen to unambiguously express the gender of the occupational
noun, \citep{Renduchintala21investigating} also found that a generic multilingual
system translating out of English made more errors for feminine than for
masculine nouns, a trend that is observed in 20~languages.

Noting the limitations of artificial datasets, \citep{Gonen20automatically}
develop a methodology to mine actual instances of  likely biased translations in
large corpora: these are found by automatically generating  minimal contrasts in
English source (e.g.\ replacing one noun by another)  yielding a gender change in
the target sentence. For instance, replacing 'doctor' by 'nurse' in the English
might trigger a gender change in the corresponding translation in Russian.

Other studies also investigated the influence of socio-professional
parameters such as profession types and the importance and the
correlation with qualifying adjectives. Using the European
multilingual classification of Skills, Competences and Occupations
(ESCO) data, \newcite{Marzi21latraduction} suggests insufficient biodiversity of
the data in the training sets of neural translation systems. Focusing
on 73~hypernyms from the 2,942 ESCO occupational nouns, she evidenced
a gender gap by comparing the translations from Google Translate,
DeepL and Microsoft Translator in the two directions for the
French/Italian language pair. She built a dataset with respectively
``competence'' (i.e.\ \textit{intelligent}) and ``appearance'' (i.e.\ 
\textit{beautiful}) adjectives (ADJ) in the following pattern <A very
\texttt{[ADJ]} \texttt{[N]} entered the room>. The data was manually
analyzed. Adjectives seem to have no influence for the translation of
masculine nouns, but competence adjectives affect the translation of
feminine nouns more severely than appearance adjectives. 

\citet{Zhao18gender} studies gender bias in ELMo embeddings
using probing techniques. In this study, biases in the embeddings also
implied biases in a pronoun reference resolution task using the
WinoGender dataset. Balancing data, and using averaged
representations, to a certain extend, helped remove this bias.

Analyzing misclassified occupations in terms of gender,
\citet{Costajussa20gender} investigated the architectural bias for the
translation of occupational nouns, suggesting that using language-specific
encoders and decoder yields less bias than a shared encoder-decoder
architecture. Considering the attention patterns in the first two decoder layers, this paper shows
that language-specific systems pay more attention to the determiner and occupational nouns, while
bilingual models seem to rely more on the determiner. In the
language-specific case, the embeddings are reported to encode more
gender information. 

Other architectural biases are considered by \citet{Renduchintala21gender}, who observed that gender bias is amplified when the system is optimised for speed. Using a dictionary of occupations for  English to Spanish and English to German, they showed that correct translation rates degrade much faster than BLEU scores when limiting the beam-size to 1 during beam search or using low-bit quantization.

Finally, another line of research focuses on \emph{mitigating} gender bias. This
can be either achieved by working on the system's internal represention \cite{Escude19equalizing},
or by creating a more balanced training data where occupational roles are
equally distributed between genders via counterfactual data augmentation
\citep{Hallmaudslay19name,Zmigrod19counterfactual}.
As discussed in \cite{Saunders20genderbias}, a cheaper, yet effective alternative to data
augmentation, is to resort to domain adaptation techniques.

\section{Discussion and Conclusion \label{sec:ccl}}

Our paper investigated the different pathways for gender transfer. We
created a dataset inspired by previous research to test several
hypotheses. Our novel contribution is that we simultaneously mobilized
several techniques, probing and manipulating. We extended the scope of
the investigation of the locus of gender transfer beyond the
determiner/noun analysis of \citet{Costajussa20gender} and questioned
the role of predicates and epicene determiners for French.
Our results show that gender information is present on the representation of all tokens built by the encoder and the decoder and suggest that the choice of the English possessive pronoun is distributed and is not based on the sole information contained in the representation of the French possessive pronoun. In our future research, we plan to identify how information is used to choose the form of the English pronoun and to generalize our observations to other languages and to other syntactic divergences.

\section*{Acknowledgements}

This work was partially funded by the NeuroViz project (Explorations and visualizations of a neural translation system), supported by the Ile-de-France Region within the DIM RFSI 2020 funding framework. This publication has emanated from research supported in part by the 2020
{\'e}mergence research project SPECTRANS, under the ANR grant (ANR-18-IDEX-0001, Financement IdEx Universit{\'e} de Paris).

%
\bibliography{neuroviz.bib}
\bibliographystyle{aclnatbib}






\end{document}